\documentclass[10pt,twocolumn,letterpaper]{article}

\usepackage[pagenumbers]{iccv} 

\usepackage{xcolor}

\definecolor{iccvblue}{rgb}{0.21,0.49,0.74}
\usepackage{xcolor}         
\usepackage{color, colortbl}
\definecolor{citecolor}{HTML}{2980b9}
\definecolor{linkcolor}{HTML}{c0392b}
\usepackage{xcolor}
\usepackage{color, colortbl}
\usepackage{multirow}
\usepackage{microtype}
\usepackage{graphicx}
\usepackage{amsmath}
\usepackage{amssymb}
\usepackage{booktabs} 
\usepackage{amssymb}
\usepackage{subcaption}
\usepackage{makecell}
\usepackage{multirow}
\usepackage{framed}
\usepackage{enumitem}
\usepackage{wrapfig}
\usepackage{diagbox}
\usepackage{adjustbox}
\usepackage{tcolorbox}
\usepackage{changepage}
\usepackage{inconsolata}
\makeatletter
  \newcommand\figcaption{\def\@captype{figure}\caption}
  \newcommand\tabcaption{\def\@captype{table}\caption}
\makeatother
\definecolor{darkorange}{HTML}{FF8C00}
\definecolor{chocolate}{HTML}{D2691E}
\definecolor{darkgreen}{HTML}{006400}
\definecolor{darkblue}{HTML}{00008B}
\definecolor{mediumblue}{HTML}{0000CD}
\definecolor{dodgerblue}{HTML}{1E90FF}
\definecolor{royalblue}{HTML}{4169E1}
\definecolor{shadecolor}{RGB}{237,237,237}
\definecolor{backred}{RGB}{255, 190, 190}
\definecolor{backblue}{RGB}{208, 230, 251}
\definecolor{backblue}{RGB}{210, 230, 250}
\usepackage{array}
\newcommand\blfootnote[1]{%
  \begingroup
  \renewcommand\thefootnote{}\footnote{#1}%
  \addtocounter{footnote}{-1}%
  \endgroup
}
\newcolumntype{C}[1]{>{\centering\arraybackslash}m{#1}}
\usepackage[pagebackref,breaklinks,colorlinks,allcolors=iccvblue]{hyperref}
\newcommand{\method}{\textsc{StyleMotif}\xspace} 
\def\paperID{2262} 
\def\confName{ICCV}
\def\confYear{2025}

\title{\method: Multi-Modal Motion Stylization using Style-Content Cross Fusion}

\author{Ziyu Guo$^\dagger$\vspace{2pt}\\
CUHK, MiuLar Lab\\
{\tt\small ziyuguo@link.cuhk.edu.hk}
\and
Young Yoon Lee\vspace{2pt}\\
Roblox\\
{\tt\small ylee@roblox.com}
\and
Joseph Liu\vspace{2pt}\\
Roblox\\
{\tt\small josephliu@roblox.com}
\and
Yizhak Ben-Shabat\vspace{2pt}\\
Roblox\\
{\tt\small ibenshabat@roblox.com}
\and
Victor Zordan\vspace{2pt}\\
Roblox\\
{\tt\small vbzordan@roblox.com}
\and
Mubbasir Kapadia\vspace{2pt}\\
Roblox\\
{\tt\small mkapadia@roblox.com}
\and
\mbox{}
\vspace{8pt}
\centerline{Project Page: \url{https://stylemotif.github.io}}
}

\begin{document}
\maketitle

\begin{abstract}
We present \textbf{\method}, a novel Stylized Motion Latent Diffusion model, generating motion conditioned on both content and style from multiple modalities. Unlike existing approaches that either focus on generating diverse motion content or transferring style from sequences, \method seamlessly synthesizes motion across a wide range of content while incorporating stylistic cues from \textbf{multi-modal} inputs, including motion, text, image, video, and audio. To achieve this, we introduce a style-content cross fusion mechanism and align a style encoder with a pre-trained multi-modal model, ensuring that the generated motion accurately captures the reference style while preserving realism. Extensive experiments demonstrate that our framework surpasses existing methods in stylized motion generation and exhibits emergent capabilities for multi-modal motion stylization, enabling more nuanced motion synthesis. 
Source code and pre-trained models will be released upon acceptance. 
\end{abstract}    

\blfootnote{$^\dagger$ Work done as an intern at Roblox.}
\section{Introduction}
\label{sec:intro}
Human motion generation is a fundamental task in computer graphics and animation, enabling the synthesis of realistic and expressive human movements. Broadly, human motion can be characterized by two complementary aspects: \textit{content}, which defines the underlying action (\eg, walking, jumping), and \textit{style}, which encodes variations such as personal flair, emotional expression, or cultural influences (\eg, jubilant, aggressive). This separation allows for greater control and flexibility in generating motion, making it particularly valuable in creative industries like game development, film production, and virtual reality. However, traditional approaches to stylized motion generation often depend on manual processes such as motion capture or keyframe animation, which are costly, time-consuming, and labor-intensive.

\begin{figure}[t!]
\centering
\includegraphics[width=0.48\textwidth]{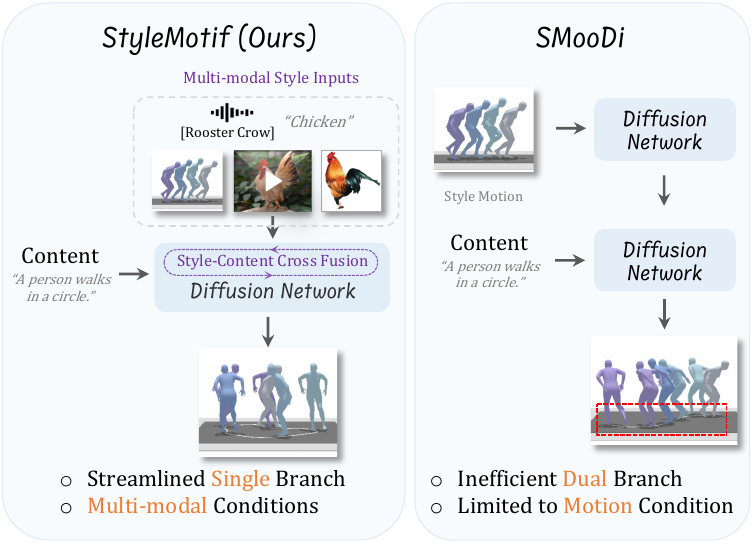}
\caption{\textbf{Comparison of Our Proposed \method Framework with SMooDi.} Unlike SMooDi’s dual-branch design, which increases model complexity and training overhead, \method employs a streamlined single-branch structure, enabling efficient multi-modal motion stylization while preserving motion realism.} 
\label{teaser}
\end{figure}

Recent progress in text-to-motion (T2M) diffusion frameworks~\cite{tevet2023human,chen2023executing,karunratanakul2023gmd} has greatly advanced the ability to translate natural-language prompts into realistic human motions. By leveraging powerful denoising diffusion models, these approaches capture intricate spatiotemporal dependencies in the data, enabling coherent sequences of human movements to be generated directly from brief textual descriptions. Despite their success in content fidelity and diversity, most current T2M diffusion methods concentrate primarily on \textit{what} action is performed, while overlooking \textit{how} it is performed, namely, its stylistic details. Simply appending a separate style-transfer module to text-driven motion diffusion pipelines can introduce additional complexity and risk of compounding errors in the final output.

In parallel, motion style transfer has been actively studied to infuse stylistic cues from a reference motion (or style data) into another sequence~\cite{aberman2020unpaired,jang2022motion,mason2022local,tao2022style}.
Although many of these methods effectively disentangle content and style for small-scale tasks, the pipeline becomes cumbersome when a large variety of content motions need to be stylized. Moreover, they commonly assume that the input or target sequences are high-quality motion data. 
In scenarios where content motions are synthetically generated, or partially noisy, the transfer process can deteriorate, leading to undesirable motion artifacts or compromised style fidelity.

To address the need for simultaneously controlling both content and style, some recent works have merged style encoding with diffusion-based motion generation. 
Among these, the most recent and representative approach~\cite{zhong2025smoodi} augments a pre-trained latent diffusion model~\cite{chen2023executing} with a style adaptor and classifier-based style guidance, achieving stylized motion from textual prompts and motion-style references. While effective, this method relies on additional training branches, which shares structural similarities with ControlNet~\cite{zhang2023adding} as shown in Figure~\ref{teaser}, which increases model complexity and training overhead. It is also constrained to motion as the primary style input. Concurrent work~\cite{li2024mulsmo} proposes a bidirectional control flow mechanism to mitigate conflicts between style and content, extending style control to multiple modalities. However, this approach also adopts a dual-branch design with substantial training overhead and limited applicability across modalities.

To this end, we propose \textbf{\method}, a new framework for \textit{multi-modal motion stylization} that unifies text-to-motion diffusion with style conditioning in a \textit{single}-branch structure. Specifically, we leverage a pre-trained motion latent diffusion model (MLD)~\cite{chen2023executing} to preserve strong content generation capabilities, and seamlessly integrate \textit{style features} extracted from a dedicated encoder, which is aligned with a multi-modal foundation model~\cite{girdhar2023imagebind}. In contrast to previous works, \method avoids duplicating large portions of the network or relying on specialized style branches. Instead, we introduce a \textit{style-content cross fusion} mechanism, which injects stylistic cues into the diffusion process while maintaining motion realism. As a result, our \method not only yields more robust stylized outputs but also supports diverse style signals, such as motion, text, images, audio, or video clips, via the alignment in multi-modal feature space.

\noindent We summarize our main contributions as follows:
\begin{itemize}
   \item We present \textbf{\method}, a stylized latent motion diffusion framework that unifies diverse motion content and \textit{multi-modal} styles within a compact \textit{single}-branch design. 
   \item We propose a \textit{style-content cross fusion module} that injects stylistic cues into the diffusion denoising process, achieving faithful stylization without compromising motion realism and ensuring efficiency. 
   \item We achieve a unified multi-modal style feature space and unveil new \textit{emergent capabilities} through \textit{multi-modal alignment}, which accommodate various sources including motion, text, images, audio, and video, for flexible and versatile multi-modal style control.
   \item Extensive experiments demonstrate that \method consistently outperforms existing methods regarding style expressiveness, content preservation, and efficiency.
\end{itemize}

\section{Related Work}
\label{sec:related}
\paragraph{Human Motion Generation.}
Recent progress in human motion generation~\cite{Guo_2022_CVPR,guo2024momask,pinyoanuntapong2024mmm,pinyoanuntapong2024bamm,wan2023tlcontrol,wu2024thor,petrovich2024multi,wu2024motionllm,chen2024motionllm,dai2024motionlcm,raab2024monkey,xu2024interdreamer,cen2024generating} has been driven by transformer~\cite{li2024lamp, guo2024momask} and diffusion models~\cite{rombach2022high,zhang2023adding,dai2024motionlcm,andreou2024lead}, showing great potential in producing realistic and diverse motions. 
These approaches have shown great potential in producing realistic and diverse motion sequences. 
For example, Momask~\cite{guo2024momask} improves motion generation using a residual VQ-VAE. Similarly, LaMP~\cite{li2024lamp} introduces a motion-aware text encoder and a motion-to-text language model to enhance motion quality through text conditioning. Diffusion models, in particular, have become a key approach in motion generation~\cite{tevet2023human,chen2023executing,zhang2024motiondiffuse,yuan2023physdiff,karunratanakul2023gmd,rempeluo2023tracepace,huang2023diffusion,xu2023interdiff}. MDM~\cite{tevet2023human} introduces a motion diffusion model that operates directly on raw motion data to capture the relationship between motions and input text conditions. MLD~\cite{chen2023executing} improve efficiency by embedding the diffusion process in latent space, reducing computational cost. They also allow conditioning on specific constraints, such as predefined trajectories~\cite{xie2023omnicontrol,karunratanakul2023gmd,wan2023tlcontrol} or human-object interactions~\cite{peng2023hoi,wu2024thor,christen2024diffh2o}, enabling greater control and diversity. 
Our work leverages pre-trained motion latent diffusion model~\cite{chen2023executing} and multi-modal foundation models~\cite{girdhar2023imagebind} to achieve stylized human motion generation while maintaining high motion quality.

\begin{figure*}[t]
\centering
\includegraphics[width=\textwidth]{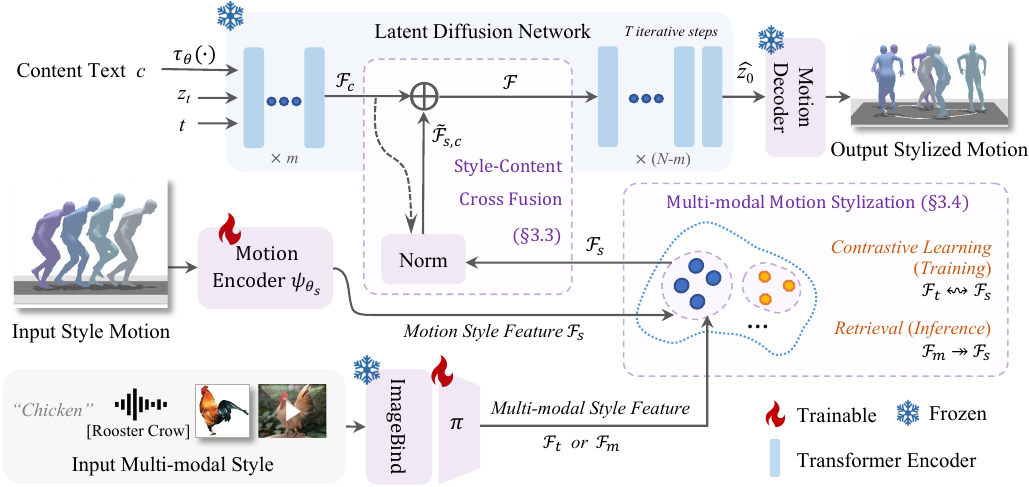}
\caption{\textbf{Overall Pipeline of \method}, a single diffusion branch framework for multi-modal motion stylization.  Given a text prompt and a reference style from various modalities, our model extract style features and fuse them with content by style-content cross fusion. Through multi-modal alignment with contrastive learning, we enable seamless multi-modal conditioning and flexible stylization across motion, text, images, audio, and video.}
\label{pipeline}
\end{figure*}

\paragraph{Motion Stylization.}
Motion stylization~\cite{aberman2020unpaired,park2021diverse,jang2022motion,tao2022style,mason2022local,song2024arbitrary,raab2023single,kim2024most,qian2024smcd,tang2023rsmt,wen2021autoregressive,xu2020hierarchical} involves transferring stylistic features from a reference motion to a source motion/textual prompt, enabling creative transformations while preserving the original motion content. Early methods, like those in motion style transfer~\cite{aberman2020unpaired,jang2022motion}, typically separate motion content and style for recombination. For instance, Aberman et al.~\cite{aberman2020unpaired} used a generative adversarial network to decouple style from content without paired data, while Motion Puzzle~\cite{jang2022motion} allows style control for individual body parts, and Guo et al.~\cite{guo2024generative} utilized pretrained motion models for better style integration. Recent approaches, like SMooDi~\cite{zhong2025smoodi}, generate stylized motion from text and style sequences using style guidance. However, these models face two main challenges: \textit{(1)} \textit{parallel-inefficient dual-branch frameworks}, and (\textit{2)} \textit{limited to style motion for guidance}. 
Another concurrent approach, MulSMo~\cite{li2024mulsmo}, attempts to address some of these by introducing a bidirectional control flow between style and content networks, reducing conflicts. Still, it remains restricted to a few modalities and relies on dual branches. 
In this work we eliminate the need for a dual-branch framework, resulting in a simpler, more effective, and efficient approach to stylized motion generation.
Our method also enables efficient multi-modal motion stylization, supporting text, image and multi-modal inputs while maintaining both high stylization quality and motion realism.

\paragraph{Multi-modality Learning.} Recent advancements in multi-modality learning~\cite{desai2021virtex,fang2021clip2video,nagrani2022learning} have revolutionized various domains by enabling joint understanding across diverse data types. Foundational models like CLIP~\cite{radford2021learning} and its extension~\cite{zhang2022tip,zhang2022pointclip,guo2022calip,zhu2023pointclip} establish robust vision-language alignment, while ImageBind~\cite{girdhar2023imagebind}, Point-Bind~\cite{guo2023point}, and Language-Bind~\cite{zhu2023languagebind} expand this paradigm to diverse modalities, demonstrating emergent cross-modal capabilities. The rise of multi-modal large language models~\cite{li2024llava,han2023imagebind,chen2024llm,zhang2024mavis,yang2023lidar,li2024manipllm,guo2025sciverse} further enhances semantic understanding through unified text-visual processing, achieving superior performance across 2D images~\cite{zhang2024llama,gao2023llama,li2408llava}, 3D point clouds~\cite{guo2023point,guo2025pisa,tang2025exploring}, and complex reasoning scenarios~\cite{guo2025can,zhang2024mathverse,jiang2025mme}. For human motion, some approaches~\cite{chen2024motionllm,jiang2023motiongpt,guo2024momask} incorporate multiple modalities into human motion understanding and generation tasks. Our work introduces the first framework for multi-modal guided motion stylization, achieving seamless style injection from diverse modalities through a single-branch diffusion architecture with style-content cross fusion.

\section{\method: Multi-Modal Motion Stylization with Style-Content Cross Fusion}
\label{sec:method}
We propose \method, a novel framework for stylized motion synthesis that combines style-content cross fusion and multi-modal motion stylization, as illustrated in Figure~\ref{pipeline}.
Our approach integrates a style-content cross fusion mechanism that allows for coherent feature blending, ensuring the generated motion accurately reflects the reference style while maintaining the content's realism (\S\ref{s3.3}).
This mechanism is complemented by a multi-modal motion stylization strategy, which leverages inputs from diverse modalities such as image, video, audio, and text to provide control over the stylization process (\S\ref{s3.4}). This framework enables highly flexible and realistic stylized motion synthesis, which offers greater control and diversity.

\subsection{Overview}
\label{s3.1}

\paragraph{Preliminaries.} 
Motion Latent Diffusion (MLD)~\cite{chen2023executing} formulates a conditional latent diffusion model by training $\epsilon_{\theta}(z_t,t,c)$ to denoise a sequence of latents $\{z_t\}^{T}_{t=0}$, where $z_t \in \mathbb{R}^{n \times d}$ represents the motion latent at timestep $t$, conditioned on the distribution $p(z_t|c)$. A conditional domain encoder $\tau_{\theta}(c)$, such as CLIP~\cite{radford2021learning}, enables text-to-motion tasks, and the model is trained as $\epsilon_{\theta}(z_t,t,\tau_{\theta}(c))$. To incorporate additional style conditioning, a style condition $s$ can be introduced with its own encoder $\psi_{\theta}(s)$, which may take the form of a motion or text encoder. This extends the model to $\epsilon_{\theta}(z_t,t,\tau_{\theta}(c),\psi_{\theta_s}(s))$. 
SMooDi~\cite{zhong2025smoodi} adopts a ControlNet~\cite{zhang2023adding}-style approach for style conditioning by creating a trainable copy of the neural network weights $\theta_s$ from the original MLD model $\theta_c$. This copy includes zero-initialized linear layers, denoted as $\mathcal{Z}(\cdot;\cdot)$. The output $\mathcal{F}^{i}(z_t,t,\tau_{\theta}(c),\psi_{\theta}(s);{\theta_c})$ of the $i$-th MLD block, now conditioned on style, is computed as:
\begin{align}
\mathcal{F}^{i}(z_t,t,\tau_{\theta}(c),&\psi_{\theta_s}(s);{\theta_c}) = \mathcal{F}^{i}(z_t,t,\tau_{\theta}(c);{\theta_c}) \notag\\
&+ \mathcal{Z}(\mathcal{F}^{i}(z_t,t,\tau_{\theta}(c),\psi_{\theta_s}(s);{\theta_{s}});\theta_{z_i})
\end{align}
A key property of this formulation is that since $\theta_{z_i}$ is initialized to zero, thus $\mathcal{F}^{i}(z_t,t,\tau_{\theta}(c), \psi_{\theta}(s)) = \mathcal{F}^{i}(z_t,t,\tau_{\theta}(c))$ at the beginning. However, the tradeoff of that method is that it needs to maintain the additional parameters $\theta_s$ and $\theta_{z_i}$ in order for style transfer.

\paragraph{\method.} We propose \textbf{\method}, eschewing zero linear layers in favor of injecting it directly via statistical manipulation, with the pipeline shown in Figure~\ref{pipeline} describing our approach. Our \method utilizes latent space diffusion within a single generative branch, building upon a pretrained MLD model~\cite{chen2023executing}. 
Instead of perturbing the outputs of each block $i$ via zero initialized layer $\mathcal{Z}(\mathcal{F}^{i}(z_t,t,\tau_{\theta}(c),\psi_{\theta}(s);{\theta_{s}});\theta_{z_i})$, \method replaces this with a statistically transformed style embedding that is injected into the original MLD branch. This simplifies the model while ensuring high-quality stylization results, whose details are presented in \S~\ref{s3.3}. Also, we follow SMooDi’s generation guidance and training scheme to ensure high-quality stylized motion synthesis. We use a hybrid guidance strategy that balances content fidelity and style adherence by combining classifier-free and classifier-based techniques during diffusion sampling. Implementation details of the learning scheme are provided in the Supplementary Material.

\subsection{Style-Content Cross Fusion}
\label{s3.3}
\paragraph{Style Encoder Pre-training.}
To establish a robust foundation for style-content fusion, we combine the content knowledge from the pre-trained MLD's~\cite{chen2023executing} VAE with the style knowledge from the 100STYLE~\cite{mason2022local} dataset. The MLD's VAE is pre-trained on the HumanML3D~\cite{Guo_2022_CVPR} dataset, which provides extensive understanding of content motion. Building on this, we further fine-tune the model on the 100STYLE dataset in a variational autoencoding manner to initially align the content and style data distributions in the latent space. After training, we discard the decoder and retain only the encoder as the motion style encoder. This reconstruction task enables the encoder to learn robust motion feature representations, which are essential for supporting the subsequent stylization process. By integrating content and style knowledge in this way, we ensure a strong foundation for seamless style-content fusion.

\paragraph{Style-Content Cross Normalization.} 
To train a stylized diffusion model $\epsilon_{\theta}(z_t,t,\tau_{\theta}(c),\psi_{\theta}(s))$, we effectively fuse content and style features directly within the latent space diffusion process \textit{instead of} using dual-branch networks or separate control mechanisms~\cite{zhang2023adding}. 
Given the output features $\mathcal{F}^i$ of the $i$-th block, we derive the content features $\mathcal{F}_c^i=\mathcal{F}^{i}(z_t,t,\tau_{\theta}(c);{\theta_c})$ from the input text $c$, and the style features $\mathcal{F}_s=\psi_{\theta_s}(s)$ are extracted from the reference motion sequence $s$ using the pre-trained style encoder. 
 To perform the fusion, we first compute the mean $\mu_c$ and variance $\sigma_c^2$ of the content features across the feature dimension:

\begin{align}
\mu_c &= \frac{1}{D} \sum_{j=1}^{D} \mathcal{F}^{i,j}_c, \\
\sigma_c^2 &= \frac{1}{D} \sum_{j=1}^{D} \left( \mathcal{F}^{i,j}_c - \mu_c \right)^2,
\end{align}

\noindent where $\mathcal{F}^{i,j}_c$ denotes the content features of the $i$-th block and the $j$-th feature element.
Next, we normalize the \textit{style} features $\mathcal{F}_s$ using these \textit{content} statistics, ensuring that the style features are adapted to the content's statistical properties:
\begin{align}
\widetilde{\mathcal{F}}_{s,c} &= \frac{\mathcal{F}_s - \mu_c}{\sqrt{{\sigma_c}^2 + \eta}},
\end{align}
\noindent where $\eta$ is a constant added for numerical stability. This style-content cross normalization ensures that the style features are smoothly integrated with the content features while maintaining the content's original structure.
After that, we add the normalized style features to the content features, formulating our final cross-normalization as
\begin{align}
\mathcal{F}^{i}(z_t,t,\tau_{\theta}(c),&\psi_{\theta_s}(s);{\theta_c}) = \mathcal{F}_c^i  +\gamma  \cdot \widetilde{\mathcal{F}}_{s,c}, 
\label{gamma}
\end{align}
where $\gamma$ is a parameter used for scaling the normalized value. $\widetilde{\mathcal{F}}_{s,c}$ can be thought of as a perturbation of $\mathcal{F}_c^i$ as it is scaled within its range. Notably, the fusion process is performed only once after the $m$-th block during the denoising process to avoid distorting the content while effectively introducing the style. This efficient fusion method eliminates the need for additional learnable parameters, as it is based solely on statistical transformations, achieving high-quality results with minimal computational overhead.

\subsection{Multi-Modal Motion Stylization}
\label{s3.4}
Our \method extends to support multi-modal motion stylization. To achieve this, we integrate a pre-trained multi-modal foundation model~\cite{girdhar2023imagebind}, which provides a unified, multi-modal aligned feature space, enabling effective cross-modal alignment between motion and other modalities. By leveraging this alignment, our model can flexibly combine multiple input modalities, such as text, image, audio, and video, to guide the stylization process in a comprehensive manner. This capability results in \textit{\textbf{emergent abilities}} for multi-modal motion stylization, allowing for more nuanced outputs.

\begin{figure*}[t!]
\centering
\begin{minipage}[c]{0.59\textwidth}
\vspace{2pt}
\centering
\begin{adjustbox}{width=\linewidth}
\begin{tabular}{l|cccccc}
\toprule
\small
\multirow{2}*{\makecell*[l]{\large Method}}
& \multirow{2}{*}{SRA\ $\uparrow$} & \multirow{2}{*}{FID\ $\downarrow$}&\multirow{2}{*}{MM Dist\ $\downarrow$}& R-Precision\ $\uparrow$ &{Diversity}&Foot Skate\\
&&&&(Top-3) &$\rightarrow$ &Ratio $\downarrow$\\
\cmidrule{1-7}
\rowcolor{gray!8}\multicolumn{7}{c}{\textit{\textbf{Motion}-Guided Stylization}}\\
\cmidrule{1-7}
MLD + Aberman~\cite{aberman2020unpaired}&54.37&3.309&5.983&0.406&8.816&0.347\\
MLD + Motion Puzzle~\cite{jang2022motion}&63.77&6.127&6.467&0.290&6.476&0.185\\
SMooDi~\cite{zhong2025smoodi} &72.42&1.609&4.477&0.571&9.235&0.124\\
\textbf{\method (Ours)} &\textbf{77.65}&\textbf{1.551}&\textbf{4.354}&\textbf{0.586}&7.567 &\textbf{0.097}\\
\cmidrule{1-7}
\rowcolor{gray!8}\multicolumn{7}{c}{\textit{\textbf{Text}-Guided Stylization}}\\
\cmidrule{1-7}
MLD + ChatGPT~\cite{zhong2025smoodi} & 4.82 & 0.614 & 4.313 & 0.605 & 8.836 & 0.131\\
\textbf{\method (Ours)} & \textbf{56.71} & \textbf{0.603} & \textbf{3.684} & \textbf{0.690} & 9.101 &\textbf{0.101}\\
\bottomrule
\end{tabular}
\end{adjustbox}
\tabcaption{\textbf{Quantitative Results for Motion-Guided and Text-Guided Stylization.} \textbf{Bold} values denote the best performance. As there is no ground-truth reference for Diversity, no value is highlighted in bold; but the metric is provided for reference.}
\label{stylization}
\end{minipage}\quad
\begin{minipage}[c]{0.37\textwidth}
\small
\centering
\vspace{2pt}
\begin{adjustbox}{width=0.95\linewidth}
\begin{tabular}{lcccc}
\toprule
\multirow{2}{*}{Method}&\multirow{2}{*}{SRA\ $\uparrow$}&\multirow{2}{*}{FID\ $\downarrow$}&Foot Skate\\
&&&Ratio $\downarrow$\\
\midrule
MLD + Aberman ~\cite{aberman2020unpaired}&61.01&3.892&0.338\\
MLD + Motion Puzzle ~\cite{jang2022motion}&67.23&6.871&0.197\\
SMooDi ~\cite{zhong2025smoodi}&65.15&1.582&0.095\\
\textbf{\method (Ours)} &\textbf{68.81} &\textbf{1.375} &\textbf{0.094}\\
\bottomrule
\end{tabular}
\end{adjustbox}
\tabcaption{\textbf{Quantitative Results of Motion Style Transfer on HumanML3D~\cite{Guo_2022_CVPR} dataset.} Our method outperforms previous works in all metrics, which demonstrates effective style-content fusion for high-quality motion style transfer, providing significant advantages for downstream tasks besides motion stylization.}
\label{transfer}
\end{minipage}
\end{figure*}

\paragraph{Motion-Text Pair Curation.} To align motion with other modalities, we process a set of motion-text pairs using the curated 100STYLE subset~\cite{mason2022local}, which is carefully selected to avoid conflicts between content and style motions, so that the model can effectively learn the relationships between motion and text.
Each motion sequence is paired with a corresponding single textual label, which serves as the text prompt. The curated motion-text pairs will be used for the following cross-model alignment. Please refer to the Supplementary Material for more details.

\paragraph{Multi-Modal Alignment.} To maintain the alignment within the multi-modal space, we freeze the text encoder of ImageBind~\cite{girdhar2023imagebind} and introduce a lightweight projection layer after the encoder. This projection layer aligns the feature dimensions of the text and motion encoders. Since the multi-modal model represents each modality with a global feature, our alignment process focuses on these global features to achieve robust alignment between text and motion representations, formalized as: 
\begin{align}
    &\mathcal{F}_{t} = \pi\big(\mathcal{E}_{\text{text}}(l)\big),\\
    &\mathcal{F}_{s} = \psi_{\theta_s}(s),
\end{align}
where $l$ and $s$ are the paired input text label and style motion, \(\mathcal{F}_{t}\) and \(\mathcal{F}_{s}\) denote the feature representations from the text and style motion encoders, $\mathcal{E}_{\text{text}}$ and $\psi_{\theta_s}$, respectively, and $\pi$ represents the projection operation that maps text features into the motion feature space.  
We employ a contrastive learning loss~\cite{xue2023ulip} to align the feature spaces of motion and text, bringing them closer together in the shared multi-modal space. Thus, we obtain a unified space between motion and all the modalities. We formulated it as
\begin{align}
 \mathcal{L}_{\text{align}} =
   -\frac{1}{2}
   \sum_{(i,j)}\log\frac{\exp\frac{\mathcal{F}^{i}_{t}\cdot \mathcal{F}^{j}_{s}}{\tau_0}}{\sum_k \exp\frac{\mathcal{F}^{i}_{t}\cdot \mathcal{F}^{k}_{s}}{\tau_0}} 
    +\log\frac{\exp\frac{\mathcal{F}^{i}_{t}\cdot \mathcal{F}^{j}_{s}}{\tau_0}}{\sum_k \exp\frac{\mathcal{F}^{k}_{t}\cdot \mathcal{F}^{j}_{s}}{\tau_0}},
\label{tau0}
\end{align}
where $t$ and $s$ represent two modalities (text and style motion) and $(i, j)$ indicates a positive pair in each training batch, $k$ indexes all samples in the batch, including both positive and negative ones, and $\tau_0$ is a temperature parameter. During inference, we obtain the multi-modal (text,  image, video, or audio) features from the multi-modal input $m$, 
\begin{align}
    &\mathcal{F}_{m} = \mathcal{E}_{\text{ImageBind}}(m),
\end{align}
where $\mathcal{E}_{\text{ImageBind}}$ denotes ImageBind~\cite{girdhar2023imagebind}. We use the multi-modal features $\mathcal{F}_m$ to retrieve the most semantically similar motion features from the unified multi-modal space. The retrieved motion features are then used to guide the stylization process, ensuring that fine-grained style details are preserved and accurately reflected in the final output. This approach enhances the model's ability to generate stylized motions that are both contextually relevant and visually coherent.

\vspace{0.1cm}
\section{Experiment}
\noindent We provide extensive quantitative and qualitative analyses across multiple tasks in this section, with additional videos available in the Supplementary Material. 

\label{sec:experiment}
\vspace{0.1cm}
\subsection{Experimental Settings}
\paragraph{Implementation Details.}
We adopt the pre-trained MLD~\cite{chen2023executing} as the foundation for motion generation. The style encoder of our model derives from the encoder of MLD's VAE, 
while the projection layer after text encoder is a single Linear layer. During diffusion training, we only enable the style encoder to be trainable while freezing other parameters. The model is optimized using the AdamW optimizer~\cite{loshchilov2017decoupled} with a constant learning rate of $10^{-5}$. More implementation details are presented in the Supplementary Material.
 
\paragraph{Dataset Settings.}
We use the HumanML3D~\cite{Guo_2022_CVPR} dataset as our primary motion content dataset, which collects 14,616 motion sequences from AMASS~\cite{mahmood2019amass} and annotates 44,970 sequence-level textual description. 
To train the style network, we utilize the 100STYLE~\cite{mason2022local} dataset, containing 45,303 style motions. We also adopt the text annotations for 100STYLE from previous work~\cite{zhong2025smoodi}, which are pseudo text descriptions generated from MotionGPT~\cite{jiang2023motiongpt}.
We utilize the consistent root-velocity motion representations from HumanML3D for both the content and style data.

\paragraph{Baselines.}
We mainly evaluate our method against SMooDi~\cite{zhong2025smoodi} on motion-guided stylization and motion style transfer tasks. We also compare our text-guided stylication with `ChatGPT+MLD' approach, which utilize ChatGPT to combine style text and content text, and functions as a straightforward text-to-motion model without control capabilities~\cite{zhong2025smoodi}. Additionally, we qualitatively compare our model with SMooDi on motion-guided stylization.

\begin{figure*}[t!]
\vspace{0.1cm}
\centering
\includegraphics[width=\textwidth]{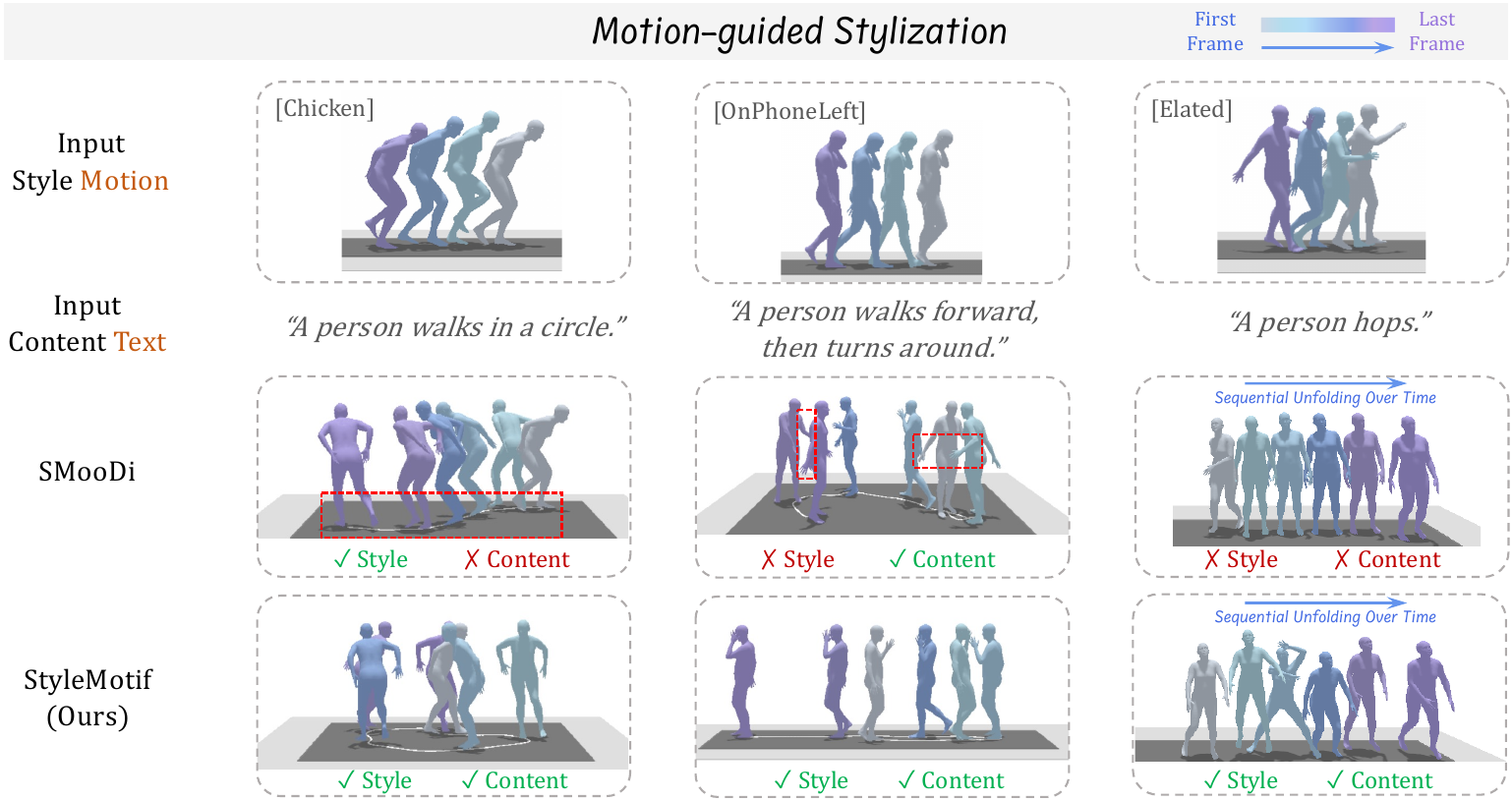}
\vspace{-0.2cm}
\caption{\textbf{Qualitative Results of Motion-Guided Stylization.} Our model generates cohesive and realistic motions that effectively align style and content, such as preserving the `circular' trajectory (first column) and `hop' content (third column). In contrast, SMooDi~\cite{zhong2025smoodi} struggles to maintain content fidelity and sometimes fails to reflect the specified style (e.g., `phone on the left' in the second colum).}
\label{motion_guided}
\end{figure*}

\vspace{0.3cm}
\paragraph{Evaluation Metrics.}
To evaluate our model, we use several metrics following previous works~\cite{zhong2025smoodi}.
We measure R-precision, Multi-modal Distance (MM Dist), Diversity, and Frechet Inception Distance (FID) to assess how well the content from text is preserved and how realistic the generated motions are.
We use Style Recognition Accuracy (SRA) to evaluate how accurately the style of the reference motion is reflected in the generated motion.
Also, we adopt the Foot Skate Ratio, which helps assess the realism of the generated motion, reducing artifacts like foot sliding. 
During evaluation, following previous works~\cite{zhong2025smoodi}, we randomly select a content text from HumanML3D~\cite{Guo_2022_CVPR} and a style motion from 100STYLE~\cite{mason2022local}, and compute SRA for the generated motion using a pre-trained classifier~\cite{zhong2025smoodi}.

\begin{figure*}[t!]
\begin{minipage}[l]{0.51\textwidth}
\centering
\includegraphics[width=\linewidth]{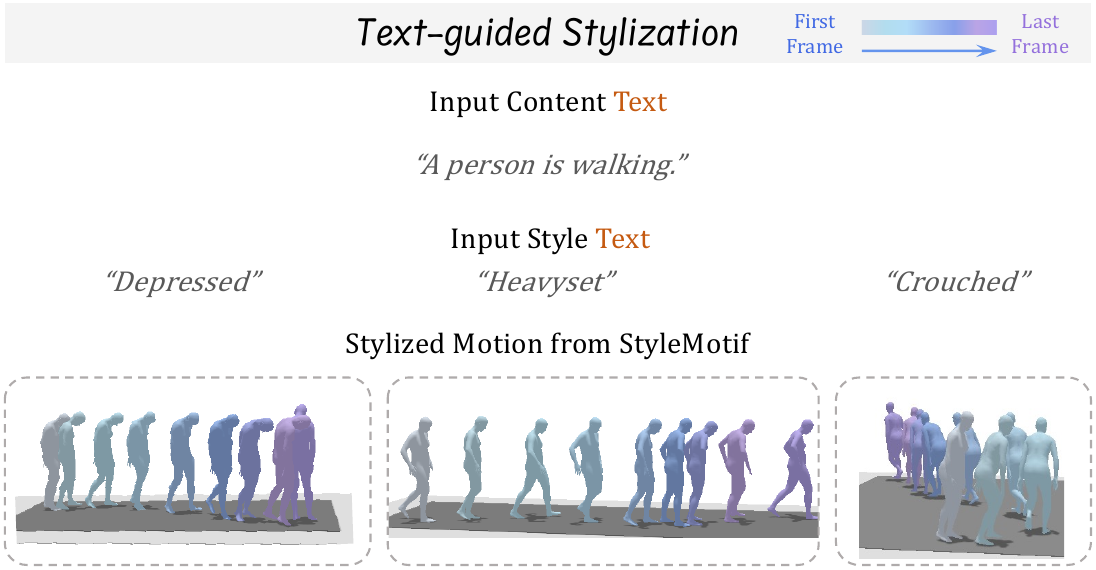}
\caption{\textbf{Qualitative Results of Text-Guided Stylization.} Our model seamlessly integrates textual style descriptions with content, producing visually coherent and stylistically consistent results. }
\label{text_guided}
\end{minipage}\hspace{5pt}
\begin{minipage}[l]{0.49\textwidth}
\includegraphics[width=\linewidth]{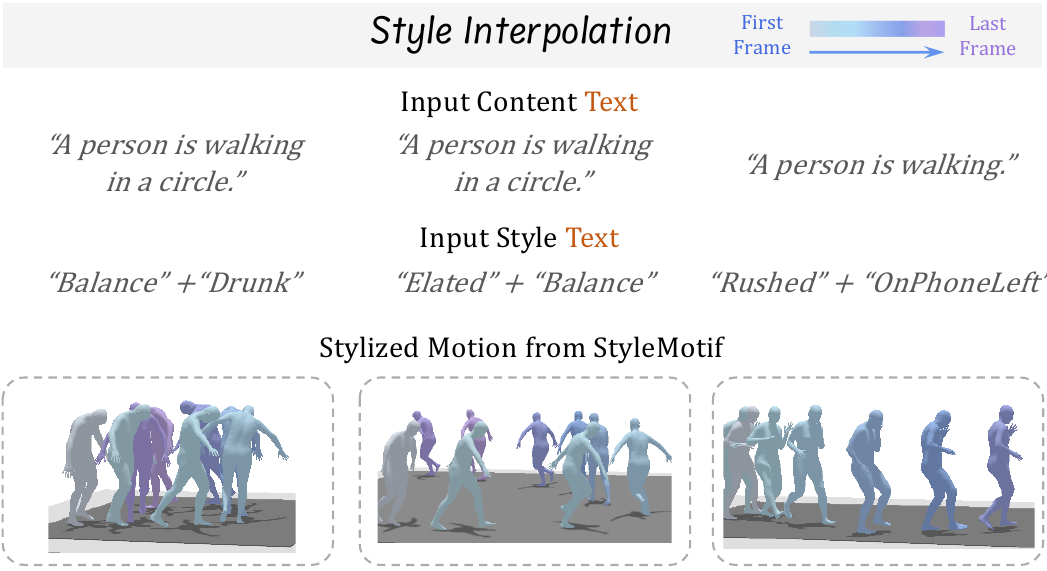}
\caption{\textbf{Qualitative Results of Style Interpolation.} Our model blends multiple style inputs while preserving content integrity, demonstrating effective style-content fusion of our model.}
\label{interpolation}
\end{minipage}
\end{figure*}

\begin{figure*}[t!]
\centering
\includegraphics[width=\textwidth]{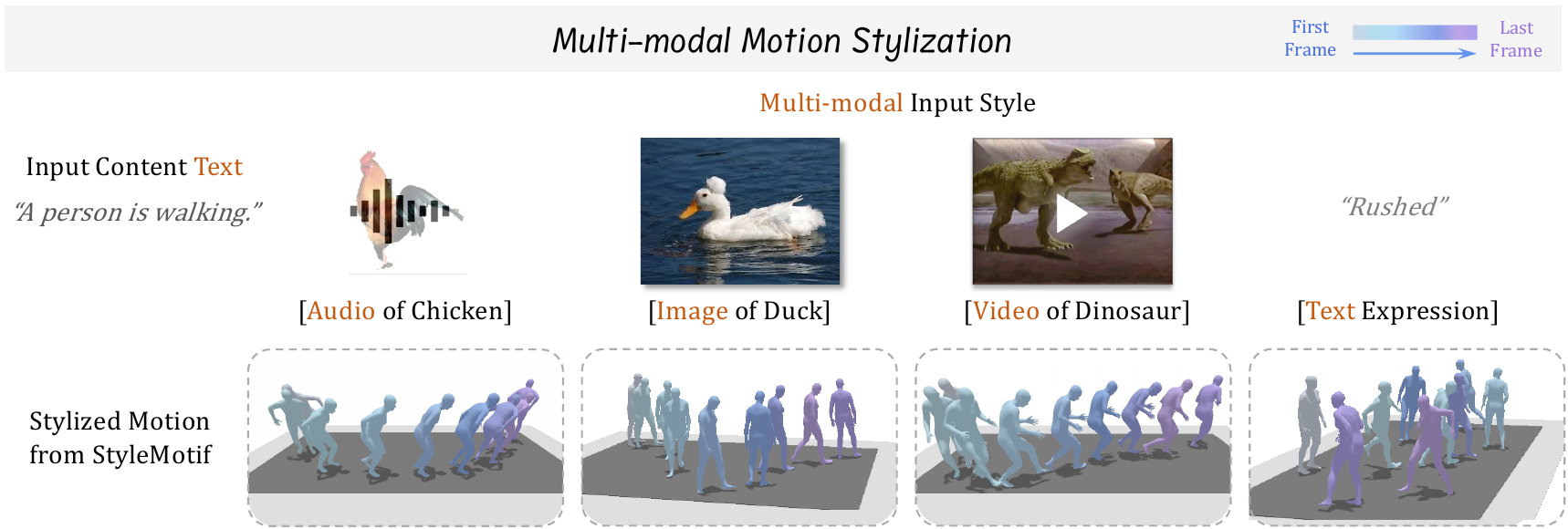}
\caption{\textbf{Qualitative Results of Multi-Modal Motion Stylization.} Our model generates stylized motions guided by diverse modalities (e.g., text, image, video, audio), effectively transferring style while preserving content integrity.}
\label{multimodal}
\end{figure*}
\subsection{Motion-guided Stylization}
\paragraph{Quantitative Analysis.}
In Table~\ref{stylization}, we report the quantitative results for motion-guided stylization, where the motion serves as style input while text prompt as content one. Our method outperforms three baseline approaches~\cite{aberman2020unpaired,jang2022motion,zhong2025smoodi} in all metrics. Specifically, we achieve a \textit{\textbf{5.23\%}} improvement in SRA while maintaining competitive performance in FID compared to the best baseline~\cite{zhong2025smoodi}. It demonstrates that our method, by effectively utilizing style-content cross fusion, generates motions that better align with the style reference while maintaining content integrity, showcasing the strength of our design in balancing style and content preservation.

\paragraph{Qualitative Analysis.}
In Figure~\ref{motion_guided}, we show some qualitative results of motion-guided stylization from our model and baseline~\cite{zhong2025smoodi}. Compared with the baseline, our model produces more cohesive stylized motions, with better alignment of style and content. For instance, in the provided examples, SMooDi fails to maintain the \textit{`circular'} trajectory or \textit{`hop'} action specified by the content or cannot accurately reflect the intended style, \textit{`phone on the left'}. This indicates that our approach more effectively integrates style and content, resulting in more realistic and consistent stylized motions.

\subsection{Text-guided Stylization}
\paragraph{Quantitative Analysis.}
For the text-guided stylization task, we use the HumanML3D~\cite{Guo_2022_CVPR} dataset for motion content and employ the single text label~\cite{mason2022local} as the style control. As reported in Table~\ref{stylization} (Bottom), our method significantly outperforms the baseline, achieving \textit{\textbf{56.71\%}} in SRA, compared to 4.82\% for `ChatGPT + MLD'. Moreover, our method maintains competitive FID, balancing style reflection with content preservation.
This indicates that our design, which combines multi-modal alignment, help utilizes the style signal in the motion-text shared spaces and achieve significant effectiveness in text-guided stylization.

\paragraph{Qualitative Analysis.}
In Figure~\ref{text_guided}, we showcase qualitative results for text-guided stylization, where our model also demonstrates strong capability in harmonizing style and content, generating high-quality and visually coherent results.

\begin{figure*}[t!]
\centering
\begin{minipage}[c]{0.26\textwidth}
\centering
\includegraphics[width=\linewidth]{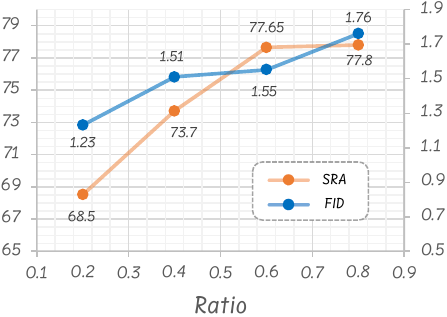}
\caption{\textbf{Ablation Study on Scaling Ratio $\gamma$ in Eq.~\ref{gamma}.} We report both SRA and FID to show the impact of the scaling ratio on both stylization and content preservation.}
\label{abla_cross}
\end{minipage}\qquad
\begin{minipage}[c]{0.67\textwidth}
\small
\centering
\begin{adjustbox}{width=\linewidth}
\begin{tabular}{C{1.5cm}C{1.5cm}|cccccc}
\toprule
\rowcolor{gray!8}\multicolumn{8}{c}{\textit{Style Encoder Pre-training Strategy}}\\
\midrule
\small
\small{HumanML3D} & \small{100STYLE}  & SRA\ $\uparrow$ & FID\ $\downarrow$ & MM Dist\ $\downarrow$ & R-Precision $\uparrow$& Diversity\ $\rightarrow$ &Foot Skate Ratio\ $\downarrow$\\
\cmidrule{1-8}
- &\checkmark &76.73 &1.788 &4.349 &0.571 &7.505 &0.101 \\
\checkmark &- &76.58 &1.635 &4.458 &0.572 &8.534 &0.109 \\
\checkmark &\checkmark &\textbf{77.65}&\textbf{1.551}&\textbf{4.354}&\textbf{0.586}&7.567 &\textbf{0.097}\\
\cmidrule{1-8}
\addlinespace[1pt]
\cmidrule{1-8}
\rowcolor{gray!8}\multicolumn{8}{c}{\textit{Text in Multi-modal Alignment}}\\
\midrule\multicolumn{2}{c|}{\makecell*[l]{Textual Expression}} & SRA\ $\uparrow$ & FID\ $\downarrow$ & MM Dist\ $\downarrow$ & R-Precision $\uparrow$& Diversity\ $\rightarrow$ &Foot Skate Ratio\ $\downarrow$\\
\cmidrule{1-8}
\multicolumn{2}{c|}{Brief Description} & 76.84 & 1.580 & 4.378 & 0.578 & 7.251 & 0.099 \\
\multicolumn{2}{c|}{Detailed Description} & 75.25 & 1.622 & 4.419 & 0.563 & 7.764 & 0.102\\
\multicolumn{2}{c|}{Single Text Label} &\textbf{77.65}&\textbf{1.551}&\textbf{4.354}&\textbf{0.586}&7.567 &\textbf{0.097} \\
\bottomrule
\end{tabular}
\end{adjustbox}
\tabcaption{\textbf{Ablation Study on Style Encoder Pre-training Strategies and Text Expression for Multi-modal Alignment.} `w. HumanML3D' and `w. 100STYLE' denote pre-training with HumanML3D~\cite{Guo_2022_CVPR} and 100STYLE~\cite{mason2022local} data respectively.}
\label{abla_}
\end{minipage}
\end{figure*}

\subsection{Motion Style Transfer}
For the motion style transfer task, we use the HumanML3D~\cite{Guo_2022_CVPR} dataset for motion content and the 100STYLE dataset~\cite{mason2022local} for motion styles. As shown in Table~\ref{transfer}, our method outperforms the baseline models across all key metrics. 
Specifically, we achieve a \textbf{\textit{3.66\%}} improvement in SRA, indicating better style reflection, and a \textbf{\textit{0.207}} reduction in FID reflecting improved realism.
These demonstrate that our framework not only improves the alignment between style and content but also provides significant advantages for downstream tasks like motion style transfer with efficient style-content cross fusion.
By maintaining content integrity while seamlessly integrating style, \method ensures more realistic and dynamically consistent stylized motions, which is critical for high-quality motion style transfer.

\subsection{Multi-Modal Motion Stylization} 
With the power of the aligned multi-modal space, our model supports stylization guided by a variety of modalities, including motion, text, image, video, audio. 
Here, we utilize the input style feature to retrieve the corresponding motion features as the style condition, while the content input is provided as text prompt.
The model then generates a stylized motion that incorporates the style from the input modality while maintaining the content integrity as specified by the text prompt.
We showcase several examples of multi-modal motion stylization in Figure~\ref{multimodal}, where different modalities guide the motion generation. For instance, when a text content \textit{``A person is walking.''} is provided alongside \textit{an image of a duck} as style input, the model retrieves a relevant motion feature and blends the content with the style of \textit{`Duckfoot'}.
As shown, the style from various inputs (text, image, video, audio) is effectively transferred to the generated motion.

\subsection{Style Interpolation}
Leveraging the aligned multi-modal space, our model enables text-guided style interpolation. Given one content text along with at least two style style texts, our model generates a motion that combines the characteristics of all input styles. When style texts are provided, we retrieve the most similar style motion features from the shared space. These features are then combined by weighted summation. The combined style features are fused with the content features using the proposed style-content cross fusion. In Figure~\ref{interpolation}, we showcase some qualitative results of style interpolation in, where the generated motions successfully blend the characteristics of both styles while maintaining the content's integrity.

\subsection{Ablation Study}
\paragraph{Style Encoder Pre-training.} 
In Table~\ref{abla_}, we investigate the impact of different pre-training strategies for the style encoder.
Our results show that pre-training on both HumanML3D~\cite{Guo_2022_CVPR} and 100STYLE~\cite{mason2022local} yields the best performance. 
Compared to pre-training solely on either one, the combined training on both datasets provides the style encoder with a richer set of prior knowledge.
This enables the model to effectively balance content preservation and style reflection, leveraging the diverse characteristics of both datasets to enhance the overall stylization quality.

\paragraph{Cross Normalization Scaling Ratio.} 
In Figure~\ref{abla_cross}, we examine the effect of different scaling ratios $\gamma$ in Eq.~\ref{gamma} on the stylization performance. As shown, the model achieves the best performance with $\gamma=0.6$, striking an optimal balance between style reflection and content preservation. The choice of scaling ratio influences both SRA and FID, where the optimal value improves stylization without sacrificing content integrity.

\paragraph{Text Expression for Multi-modal Alignment.}  
In Table~\ref{abla_}, we also explored different text representations for contrastive learning alignment. Specifically, we tested single labels (\eg, \textit{``Old"}), brief descriptions (\eg, \textit{``An old person"}), and more detailed descriptions (\eg, \textit{``An old person is moving slow and stiff"}). 
Our experiments show that using a single label produced the best results, providing a clearer and more concise style signal for the model while avoiding unnecessary complexity in the text representation.

\begin{table}[t!]
\vspace{0.2cm}
\centering
\small
\begin{tabular}{lccc}
\toprule
\multirow{2}{*}{Method} & Overall &Learnable &Inference\\
 &Parameter &Parameter &Time \\
\midrule
SMooDi ~\cite{zhong2025smoodi} &468M &13.9 M &4.0 s\\
\textbf{StyleMotif} &\textbf{462 M} &\textbf{7.8 M} &\textbf{3.1 s}\\
\textcolor{blue}{\textit{Improvement}} &\textcolor{blue}{\textit{1.3\%}} &\textcolor{blue}{\textit{43.9\%}} &\textcolor{blue}{\textit{22.5\%}} \\
\bottomrule
\end{tabular}
\vspace{-0.1cm}
\caption{\textbf{Efficiency Comparison.} For inference time, we report the average time cost (s) per sample on a single NVIDIA A100 GPU.} 
\label{efficiency}
\end{table}

\subsection{Efficiency Analysis}
We compare the efficiency of our method with SMooDi~\cite{zhong2025smoodi} in terms of learnable parameters and inference speed (seconds per sample), under the same diffusion step setting. As shown in Table~\ref{efficiency}, our model reduces the number of trainable parameters by \textit{\textbf{43.9\%}}, significantly easing training. While the overall parameter count remains comparable, our single-branch design boosts inference speed by \textit{\textbf{22.5\%}}, outperforming SMooDi’s dual-branch structure. Notably, our style encoder is deeper and thus accounts for most of the computational cost, but \textit{our single-banch design allows for highly parallelizable operations}. In contrast, SMooDi’s dual-branch approach requires output summation after each block, limiting parallel efficiency despite fewer overall parameters. Consequently, our method achieves faster practical inference and more efficient training. 

\section{Conclusion}
In this work, we introduce \textbf{\method}, a novel Stylized Motion Diffusion model capable of generating motion conditioned on both content and style from multiple modalities. Unlike prior approaches that either focus on motion generation across various content types or style transfer between sequences, \method effectively synthesizes motion while incorporating stylistic cues from \textit{multi-modal} inputs, including text, image, video, and audio. To achieve this, we introduce a \textit{style-content cross fusion} mechanism and align a style encoder with a pre-trained multi-modal model, ensuring that the generated motion accurately captures the reference style while maintaining realism. Through extensive experiments across diverse applications, we demonstrate that \method outperforms existing methods in stylized motion generation, producing high-quality, realistic results that faithfully adhere to the given style references. Moreover, our model exhibits \textit{\textbf{emergent capabilities}} for multi-modal motion stylization, enabling richer and more nuanced motion synthesis. These findings indicate the potential of \method in advancing stylized motion generation and open new avenues for future research in multi-modal-driven motion synthesis and style-aware generative models.

\section*{Limitations and Future Work}
The current limitations mainly exist in the relatively limited availability of style motion-text data, which constrains the model's ability to fully generalize across a wide variety of motion styles. 
Future work is to explore ways to unlock the potential of existing data by enhancing generalization within the current datasets, further extending the capabilities of the model to generate more diverse and complex motions, even within the constraints of limited data.


{
    \small
    \bibliographystyle{ieeenat_fullname}
    \bibliography{main}

\begin{thebibliography}{73}
\providecommand{\natexlab}[1]{#1}
\providecommand{\url}[1]{\texttt{#1}}
\expandafter\ifx\csname urlstyle\endcsname\relax
  \providecommand{\doi}[1]{doi: #1}\else
  \providecommand{\doi}{doi: \begingroup \urlstyle{rm}\Url}\fi

\bibitem[Aberman et~al.(2020)Aberman, Weng, Lischinski, Cohen-Or, and Chen]{aberman2020unpaired}
K. Aberman, Y. Weng, D. Lischinski, D. Cohen-Or, and B. Chen.
\newblock Unpaired motion style transfer from video to animation.
\newblock \emph{TOG}, 2020.

\bibitem[Andreou et~al.(2024)Andreou, Wang, Abrevaya, Cani, Chrysanthou, and Kalogeiton]{andreou2024lead}
Nefeli Andreou, Xi Wang, Victoria~Fern{\'a}ndez Abrevaya, Marie-Paule Cani, Yiorgos Chrysanthou, and Vicky Kalogeiton.
\newblock Lead: Latent realignment for human motion diffusion.
\newblock \emph{arXiv preprint arXiv:2410.14508}, 2024.

\bibitem[Cen et~al.(2024)Cen, Pi, Peng, Shen, Yang, Zhu, Bao, and Zhou]{cen2024generating}
Z. Cen, H. Pi, S. Peng, Z. Shen, M. Yang, S. Zhu, H. Bao, and X. Zhou.
\newblock Generating human motion in 3d scenes from text descriptions.
\newblock In \emph{CVPR}, 2024.

\bibitem[Chen et~al.(2024{\natexlab{a}})Chen, Du, You, Islam, Guo, Jin, Chen, and Heng]{chen2024llm}
Kexin Chen, Yuyang Du, Tao You, Mobarakol Islam, Ziyu Guo, Yueming Jin, Guangyong Chen, and Pheng-Ann Heng.
\newblock Llm-assisted multi-teacher continual learning for visual question answering in robotic surgery.
\newblock \emph{ICRA 2024}, 2024{\natexlab{a}}.

\bibitem[Chen et~al.(2024{\natexlab{b}})Chen, Lu, Zeng, Zhang, Wang, Zhang, and Zhang]{chen2024motionllm}
L.H. Chen, S. Lu, A. Zeng, H. Zhang, B. Wang, R. Zhang, and L. Zhang.
\newblock Motionllm: Understanding human behaviors from human motions and videos.
\newblock \emph{ArXiv}, 2024{\natexlab{b}}.

\bibitem[Chen et~al.(2023)Chen, Jiang, Liu, Huang, Fu, Chen, and Yu]{chen2023executing}
Xin Chen, Biao Jiang, Wen Liu, Zilong Huang, Bin Fu, Tao Chen, and Gang Yu.
\newblock Executing your commands via motion diffusion in latent space.
\newblock In \emph{CVPR}, 2023.

\bibitem[Christen et~al.(2024)Christen, Hampali, Sener, Remelli, Hodan, Sauser, Ma, and Tekin]{christen2024diffh2o}
Sammy Christen, Shreyas Hampali, Fadime Sener, Edoardo Remelli, Tomas Hodan, Eric Sauser, Shugao Ma, and Bugra Tekin.
\newblock Diffh2o: Diffusion-based synthesis of hand-object interactions from textual descriptions.
\newblock In \emph{SIGGRAPH Asia 2024 Conference Papers}, 2024.

\bibitem[Dai et~al.(2024)Dai, Chen, Wang, Liu, Dai, and Tang]{dai2024motionlcm}
Wenxun Dai, Ling-Hao Chen, Jingbo Wang, Jinpeng Liu, Bo Dai, and Yansong Tang.
\newblock Motionlcm: Real-time controllable motion generation via latent consistency model.
\newblock \emph{arXiv preprint arXiv:2404.19759}, 2024.

\bibitem[Desai and Johnson(2021)]{desai2021virtex}
Karan Desai and Justin Johnson.
\newblock Virtex: Learning visual representations from textual annotations.
\newblock In \emph{Proceedings of the IEEE/CVF conference on computer vision and pattern recognition}, pages 11162--11173, 2021.

\bibitem[Fang et~al.(2021)Fang, Xiong, Xu, and Chen]{fang2021clip2video}
Han Fang, Pengfei Xiong, Luhui Xu, and Yu Chen.
\newblock Clip2video: Mastering video-text retrieval via image clip.
\newblock \emph{arXiv preprint arXiv:2106.11097}, 2021.

\bibitem[Gao* et~al.(2023)Gao*, Han*, Zhang*, Lin, Geng, Zhou, Zhang, Lu, He, Yue, et~al.]{gao2023llama}
Peng Gao*, Jiaming Han*, Renrui Zhang*, Ziyi Lin, Shijie Geng, Aojun Zhou, Wei Zhang, Pan Lu, Conghui He, Xiangyu Yue, et~al.
\newblock Llama-adapter v2: Parameter-efficient visual instruction model.
\newblock \emph{arXiv preprint arXiv:2304.15010}, 2023.

\bibitem[Girdhar et~al.(2023)Girdhar, El-Nouby, Liu, Singh, Alwala, Joulin, and Misra]{girdhar2023imagebind}
Rohit Girdhar, Alaaeldin El-Nouby, Zhuang Liu, Mannat Singh, Kalyan~Vasudev Alwala, Armand Joulin, and Ishan Misra.
\newblock Imagebind: One embedding space to bind them all.
\newblock In \emph{Proceedings of the IEEE/CVF conference on computer vision and pattern recognition}, pages 15180--15190, 2023.

\bibitem[Guo et~al.(2022)Guo, Zou, Zuo, Wang, Ji, Li, and Cheng]{Guo_2022_CVPR}
Chuan Guo, Shihao Zou, Xinxin Zuo, Sen Wang, Wei Ji, Xingyu Li, and Li Cheng.
\newblock Generating diverse and natural 3d human motions from text.
\newblock In \emph{CVPR}, 2022.

\bibitem[Guo et~al.(2024{\natexlab{a}})Guo, Mu, Javed, Wang, and Cheng]{guo2024momask}
Chuan Guo, Yuxuan Mu, Muhammad~Gohar Javed, Sen Wang, and Li Cheng.
\newblock Momask: Generative masked modeling of 3d human motions.
\newblock In \emph{CVPR}, 2024{\natexlab{a}}.

\bibitem[Guo et~al.(2024{\natexlab{b}})Guo, Mu, Zuo, Dai, Yan, Lu, and Cheng]{guo2024generative}
Chuan Guo, Yuxuan Mu, Xinxin Zuo, Peng Dai, Youliang Yan, Juwei Lu, and Li Cheng.
\newblock Generative human motion stylization in latent space.
\newblock \emph{arXiv preprint arXiv:2401.13505}, 2024{\natexlab{b}}.

\bibitem[Guo* et~al.(2022)Guo*, Zhang*\#, Qiu, Ma, Miao, He, and Cui]{guo2022calip}
Ziyu Guo*, Renrui Zhang*\#, Longtian Qiu, Xianzheng Ma, Xupeng Miao, Xuming He, and Bin Cui.
\newblock Calip: Zero-shot enhancement of clip with parameter-free attention.
\newblock \emph{AAAI 2023 Oral}, 2022.

\bibitem[Guo et~al.(2023)Guo, Zhang, Zhu, Tang, Ma, Han, Chen, Gao, Li, Li, et~al.]{guo2023point}
Ziyu Guo, Renrui Zhang, Xiangyang Zhu, Yiwen Tang, Xianzheng Ma, Jiaming Han, Kexin Chen, Peng Gao, Xianzhi Li, Hongsheng Li, et~al.
\newblock Point-bind \& point-llm: Aligning point cloud with multi-modality for 3d understanding, generation, and instruction following.
\newblock \emph{arXiv preprint arXiv:2309.00615}, 2023.

\bibitem[Guo et~al.(2025{\natexlab{a}})Guo, Lin, Yuan, Zheng, Qiu, Jiang, Zhang, Feng, and Li]{guo2025pisa}
Zilu Guo, Hongbin Lin, Zhihao Yuan, Chaoda Zheng, Pengshuo Qiu, Dongzhi Jiang, Renrui Zhang, Chun-Mei Feng, and Zhen Li.
\newblock Pisa: A self-augmented data engine and training strategy for 3d understanding with large models.
\newblock \emph{arXiv preprint arXiv:2503.10529}, 2025{\natexlab{a}}.

\bibitem[Guo et~al.(2025{\natexlab{b}})Guo, Zhang, Chen, Gao, Jiang, Wang, and Heng]{guo2025sciverse}
Ziyu Guo, Ray Zhang, Hao Chen, Jialin Gao, Dongzhi Jiang, Jiaze Wang, and Pheng-Ann Heng.
\newblock Sciverse: Unveiling the knowledge comprehension and visual reasoning of lmms on multi-modal scientific problems.
\newblock \emph{arXiv preprint arXiv:2503.10627}, 2025{\natexlab{b}}.

\bibitem[Guo et~al.(2025{\natexlab{c}})Guo, Zhang, Tong, Zhao, Gao, Li, and Heng]{guo2025can}
Ziyu Guo, Renrui Zhang, Chengzhuo Tong, Zhizheng Zhao, Peng Gao, Hongsheng Li, and Pheng-Ann Heng.
\newblock Can we generate images with cot? let's verify and reinforce image generation step by step.
\newblock \emph{arXiv preprint arXiv:2501.13926}, 2025{\natexlab{c}}.

\bibitem[Han et~al.(2023)Han, Zhang, Shao, Gao, Xu, Xiao, Zhang, Liu, Wen, Guo, et~al.]{han2023imagebind}
Jiaming Han, Renrui Zhang, Wenqi Shao, Peng Gao, Peng Xu, Han Xiao, Kaipeng Zhang, Chris Liu, Song Wen, Ziyu Guo, et~al.
\newblock Imagebind-llm: Multi-modality instruction tuning.
\newblock \emph{arXiv preprint arXiv:2309.03905}, 2023.

\bibitem[Huang et~al.(2023)Huang, Wang, Li, Jia, Liu, Zhu, Liang, and Zhu]{huang2023diffusion}
Siyuan Huang, Zan Wang, Puhao Li, Baoxiong Jia, Tengyu Liu, Yixin Zhu, Wei Liang, and Song-Chun Zhu.
\newblock Diffusion-based generation, optimization, and planning in 3d scenes.
\newblock In \emph{CVPR}, 2023.

\bibitem[Jang et~al.(2022)Jang, Park, and Lee]{jang2022motion}
Deok-Kyeong Jang, Soomin Park, and Sung-Hee Lee.
\newblock Motion puzzle: Arbitrary motion style transfer by body part.
\newblock \emph{ACM TOG}, 2022.

\bibitem[Jiang et~al.(2023)Jiang, Chen, Liu, Yu, Yu, and Chen]{jiang2023motiongpt}
Biao Jiang, Xin Chen, Wen Liu, Jingyi Yu, Gang Yu, and Tao Chen.
\newblock Motiongpt: Human motion as a foreign language.
\newblock \emph{Advances in Neural Information Processing Systems}, 36:\penalty0 20067--20079, 2023.

\bibitem[Jiang et~al.(2025)Jiang, Zhang, Guo, Li, Qi, Chen, Wang, Jin, Guo, Yan, et~al.]{jiang2025mme}
Dongzhi Jiang, Renrui Zhang, Ziyu Guo, Yanwei Li, Yu Qi, Xinyan Chen, Liuhui Wang, Jianhan Jin, Claire Guo, Shen Yan, et~al.
\newblock Mme-cot: Benchmarking chain-of-thought in large multimodal models for reasoning quality, robustness, and efficiency.
\newblock \emph{arXiv preprint arXiv:2502.09621}, 2025.

\bibitem[Karunratanakul et~al.(2023)Karunratanakul, Preechakul, Suwajanakorn, and Tang]{karunratanakul2023gmd}
Korrawe Karunratanakul, Konpat Preechakul, Supasorn Suwajanakorn, and Siyu Tang.
\newblock Gmd: Controllable human motion synthesis via guided diffusion models.
\newblock In \emph{ICCV}, 2023.

\bibitem[Kim et~al.(2024)Kim, Kim, Chang, and Choi]{kim2024most}
Boeun Kim, Jungho Kim, Hyung~Jin Chang, and Jin~Young Choi.
\newblock Most: Motion style transformer between diverse action contents.
\newblock In \emph{Proceedings of the IEEE/CVF Conference on Computer Vision and Pattern Recognition}, pages 1705--1714, 2024.

\bibitem[Li et~al.()Li, Zhang, Guo, Zhang, Li, Zhang, Zhang, Li, Liu, and Li]{li2408llava}
Bo Li, Yuanhan Zhang, Dong Guo, Renrui Zhang, Feng Li, Hao Zhang, Kaichen Zhang, Yanwei Li, Ziwei Liu, and Chunyuan Li.
\newblock Llava-onevision: Easy visual task transfer, 2024a.
\newblock \emph{URL https://arxiv. org/abs/2408.03326}.

\bibitem[Li et~al.(2024{\natexlab{a}})Li, Zhang, Zhang, Zhang, Li, Li, Ma, and Li]{li2024llava}
Feng Li, Renrui Zhang, Hao Zhang, Yuanhan Zhang, Bo Li, Wei Li, Zejun Ma, and Chunyuan Li.
\newblock Llava-next-interleave: Tackling multi-image, video, and 3d in large multimodal models.
\newblock \emph{ICLR 2025 Spotlight}, 2024{\natexlab{a}}.

\bibitem[Li et~al.(2024{\natexlab{b}})Li, Zhang, Geng, Geng, Long, Shen, Zhang, Liu, and Dong]{li2024manipllm}
Xiaoqi Li, Mingxu Zhang, Yiran Geng, Haoran Geng, Yuxing Long, Yan Shen, Renrui Zhang, Jiaming Liu, and Hao Dong.
\newblock Manipllm: Embodied multimodal large language model for object-centric robotic manipulation.
\newblock In \emph{Proceedings of the IEEE/CVF Conference on Computer Vision and Pattern Recognition}, pages 18061--18070, 2024{\natexlab{b}}.

\bibitem[Li et~al.(2024{\natexlab{c}})Li, He, Zhong, Shen, Zuo, Qiu, Dong, Yang, and Yuan]{li2024mulsmo}
Zhe Li, Yisheng He, Lei Zhong, Weichao Shen, Qi Zuo, Lingteng Qiu, Zilong Dong, Laurence~Tianruo Yang, and Weihao Yuan.
\newblock Mulsmo: Multimodal stylized motion generation by bidirectional control flow.
\newblock \emph{arXiv preprint arXiv:2412.09901}, 2024{\natexlab{c}}.

\bibitem[Li et~al.(2024{\natexlab{d}})Li, Yuan, He, Qiu, Zhu, Gu, Shen, Dong, Dong, and Yang]{li2024lamp}
Zhe Li, Weihao Yuan, Yisheng He, Lingteng Qiu, Shenhao Zhu, Xiaodong Gu, Weichao Shen, Yuan Dong, Zilong Dong, and Laurence~T Yang.
\newblock Lamp: Language-motion pretraining for motion generation, retrieval, and captioning.
\newblock \emph{arXiv preprint arXiv:2410.07093}, 2024{\natexlab{d}}.

\bibitem[Loshchilov(2017)]{loshchilov2017decoupled}
I Loshchilov.
\newblock Decoupled weight decay regularization.
\newblock \emph{arXiv preprint arXiv:1711.05101}, 2017.

\bibitem[Mahmood et~al.(2019)Mahmood, Ghorbani, Troje, Pons-Moll, and Black]{mahmood2019amass}
Naureen Mahmood, Nima Ghorbani, Nikolaus~F Troje, Gerard Pons-Moll, and Michael~J Black.
\newblock Amass: Archive of motion capture as surface shapes.
\newblock In \emph{ICCV}, 2019.

\bibitem[Mason et~al.(2022)Mason, Starke, and Komura]{mason2022local}
Ian Mason, Sebastian Starke, and Taku Komura.
\newblock Real-time style modelling of human locomotion via feature-wise transformations and local motion phases.
\newblock \emph{Proceedings of the ACM on Computer Graphics and Interactive Techniques}, 2022.

\bibitem[Nagrani et~al.(2022)Nagrani, Seo, Seybold, Hauth, Manen, Sun, and Schmid]{nagrani2022learning}
Arsha Nagrani, Paul~Hongsuck Seo, Bryan Seybold, Anja Hauth, Santiago Manen, Chen Sun, and Cordelia Schmid.
\newblock Learning audio-video modalities from image captions.
\newblock In \emph{European Conference on Computer Vision}, pages 407--426. Springer, 2022.

\bibitem[Park et~al.(2021)Park, Jang, and Lee]{park2021diverse}
Soomin Park, Deok-Kyeong Jang, and Sung-Hee Lee.
\newblock Diverse motion stylization for multiple style domains via spatial-temporal graph-based generative model.
\newblock \emph{Proceedings of the ACM on Computer Graphics and Interactive Techniques}, 2021.

\bibitem[Peng et~al.(2023)Peng, Xie, Wu, Jampani, Sun, and Jiang]{peng2023hoi}
Xiaogang Peng, Yiming Xie, Zizhao Wu, Varun Jampani, Deqing Sun, and Huaizu Jiang.
\newblock Hoi-diff: Text-driven synthesis of 3d human-object interactions using diffusion models.
\newblock \emph{arXiv preprint arXiv:2312.06553}, 2023.

\bibitem[Petrovich et~al.(2024)Petrovich, Litany, Iqbal, Black, Varol, Bin~Peng, and Rempe]{petrovich2024multi}
Mathis Petrovich, Or Litany, Umar Iqbal, Michael~J Black, Gul Varol, Xue Bin~Peng, and Davis Rempe.
\newblock Multi-track timeline control for text-driven 3d human motion generation.
\newblock In \emph{CVPR}, 2024.

\bibitem[Pinyoanuntapong et~al.(2024{\natexlab{a}})Pinyoanuntapong, Saleem, Wang, Lee, Das, and Chen]{pinyoanuntapong2024bamm}
Ekkasit Pinyoanuntapong, Muhammad~Usama Saleem, Pu Wang, Minwoo Lee, Srijan Das, and Chen Chen.
\newblock Bamm: Bidirectional autoregressive motion model.
\newblock \emph{arXiv preprint arXiv:2403.19435}, 2024{\natexlab{a}}.

\bibitem[Pinyoanuntapong et~al.(2024{\natexlab{b}})Pinyoanuntapong, Wang, Lee, and Chen]{pinyoanuntapong2024mmm}
Ekkasit Pinyoanuntapong, Pu Wang, Minwoo Lee, and Chen Chen.
\newblock Mmm: Generative masked motion model.
\newblock In \emph{CVPR}, 2024{\natexlab{b}}.

\bibitem[Qian et~al.(2024)Qian, Xiao, Wu, Yang, Li, Wang, Wang, Kou, and Zhang]{qian2024smcd}
Ziyun Qian, Zeyu Xiao, Zhenyi Wu, Dingkang Yang, Mingcheng Li, Shunli Wang, Shuaibing Wang, Dongliang Kou, and Lihua Zhang.
\newblock Smcd: High realism motion style transfer via mamba-based diffusion.
\newblock \emph{arXiv preprint arXiv:2405.02844}, 2024.

\bibitem[Raab et~al.(2023)Raab, Leibovitch, Tevet, Arar, Bermano, and Cohen-Or]{raab2023single}
Sigal Raab, Inbal Leibovitch, Guy Tevet, Moab Arar, Amit~H Bermano, and Daniel Cohen-Or.
\newblock Single motion diffusion.
\newblock \emph{arXiv preprint arXiv:2302.05905}, 2023.

\bibitem[Raab et~al.(2024)Raab, Gat, Sala, Tevet, Shalev-Arkushin, Fried, Bermano, and Cohen-Or]{raab2024monkey}
Sigal Raab, Inbar Gat, Nathan Sala, Guy Tevet, Rotem Shalev-Arkushin, Ohad Fried, Amit~H Bermano, and Daniel Cohen-Or.
\newblock Monkey see, monkey do: Harnessing self-attention in motion diffusion for zero-shot motion transfer.
\newblock \emph{arXiv preprint arXiv:2406.06508}, 2024.

\bibitem[Radford et~al.(2021)Radford, Kim, Hallacy, Ramesh, Goh, Agarwal, Sastry, Askell, Mishkin, Clark, et~al.]{radford2021learning}
Alec Radford, Jong~Wook Kim, Chris Hallacy, Aditya Ramesh, Gabriel Goh, Sandhini Agarwal, Girish Sastry, Amanda Askell, Pamela Mishkin, Jack Clark, et~al.
\newblock Learning transferable visual models from natural language supervision.
\newblock 2021.

\bibitem[Rempe et~al.(2023)Rempe, Luo, Peng, Yuan, Kitani, Kreis, Fidler, and Litany]{rempeluo2023tracepace}
Davis Rempe, Zhengyi Luo, Xue~Bin Peng, Ye Yuan, Kris Kitani, Karsten Kreis, Sanja Fidler, and Or Litany.
\newblock Trace and pace: Controllable pedestrian animation via guided trajectory diffusion.
\newblock In \emph{CVPR}, 2023.

\bibitem[Rombach et~al.(2022)Rombach, Blattmann, Lorenz, Esser, and Ommer]{rombach2022high}
Robin Rombach, Andreas Blattmann, Dominik Lorenz, Patrick Esser, and Bj{\"o}rn Ommer.
\newblock High-resolution image synthesis with latent diffusion models.
\newblock In \emph{CVPR}, 2022.

\bibitem[Song et~al.(2024)Song, Jin, Li, Chen, Hao, Hou, Li, and Qin]{song2024arbitrary}
Wenfeng Song, Xingliang Jin, Shuai Li, Chenglizhao Chen, Aimin Hao, Xia Hou, Ning Li, and Hong Qin.
\newblock Arbitrary motion style transfer with multi-condition motion latent diffusion model.
\newblock In \emph{Proceedings of the IEEE/CVF Conference on Computer Vision and Pattern Recognition}, pages 821--830, 2024.

\bibitem[Tang et~al.(2023)Tang, Wu, Wang, Hu, Gong, Liao, Li, Kou, and Jin]{tang2023rsmt}
Xiangjun Tang, Linjun Wu, He Wang, Bo Hu, Xu Gong, Yuchen Liao, Songnan Li, Qilong Kou, and Xiaogang Jin.
\newblock Rsmt: Real-time stylized motion transition for characters.
\newblock In \emph{ACM SIGGRAPH 2023 Conference Proceedings}, pages 1--10, 2023.

\bibitem[Tang et~al.(2025)Tang, Guo, Wang, Zhang, Chen, Liu, Qu, Wang, Wang, Li, et~al.]{tang2025exploring}
Yiwen Tang, Zoey Guo, Zhuhao Wang, Ray Zhang, Qizhi Chen, Junli Liu, Delin Qu, Zhigang Wang, Dong Wang, Xuelong Li, et~al.
\newblock Exploring the potential of encoder-free architectures in 3d lmms.
\newblock \emph{arXiv preprint arXiv:2502.09620}, 2025.

\bibitem[Tao et~al.(2022)Tao, Zhan, Chen, and van~de Panne]{tao2022style}
Tianxin Tao, Xiaohang Zhan, Zhongquan Chen, and Michiel van~de Panne.
\newblock Style-erd: Responsive and coherent online motion style transfer.
\newblock 2022.

\bibitem[Tevet et~al.(2023)Tevet, Raab, Gordon, Shafir, Cohen-or, and Bermano]{tevet2023human}
Guy Tevet, Sigal Raab, Brian Gordon, Yonatan Shafir, Daniel Cohen-or, and Amit~Haim Bermano.
\newblock Human motion diffusion model.
\newblock In \emph{ICLR}, 2023.

\bibitem[Wan et~al.(2023)Wan, Dou, Komura, Wang, Jayaraman, and Liu]{wan2023tlcontrol}
Weilin Wan, Zhiyang Dou, Taku Komura, Wenping Wang, Dinesh Jayaraman, and Lingjie Liu.
\newblock Tlcontrol: Trajectory and language control for human motion synthesis.
\newblock \emph{arXiv preprint arXiv:2311.17135}, 2023.

\bibitem[Wen et~al.(2021)Wen, Yang, Fu, Gao, Sun, and Liu]{wen2021autoregressive}
Yu-Hui Wen, Zhipeng Yang, Hongbo Fu, Lin Gao, Yanan Sun, and Yong-Jin Liu.
\newblock Autoregressive stylized motion synthesis with generative flow.
\newblock In \emph{CVPR}, 2021.

\bibitem[Wu et~al.(2024{\natexlab{a}})Wu, Shi, Huang, Yu, Xu, and Wang]{wu2024thor}
Qianyang Wu, Ye Shi, Xiaoshui Huang, Jingyi Yu, Lan Xu, and Jingya Wang.
\newblock Thor: Text to human-object interaction diffusion via relation intervention.
\newblock \emph{arXiv preprint arXiv:2403.11208}, 2024{\natexlab{a}}.

\bibitem[Wu et~al.(2024{\natexlab{b}})Wu, Zhao, Wang, Tai, and Tang]{wu2024motionllm}
Qi Wu, Yubo Zhao, Yifan Wang, Yu-Wing Tai, and Chi-Keung Tang.
\newblock Motionllm: Multimodal motion-language learning with large language models.
\newblock \emph{arXiv preprint arXiv:2405.17013}, 2024{\natexlab{b}}.

\bibitem[Xie et~al.(2024)Xie, Jampani, Zhong, Sun, and Jiang]{xie2023omnicontrol}
Yiming Xie, Varun Jampani, Lei Zhong, Deqing Sun, and Huaizu Jiang.
\newblock Omnicontrol: Control any joint at any time for human motion generation.
\newblock In \emph{ICLR}, 2024.

\bibitem[Xu et~al.(2020)Xu, Xu, Ni, Yang, Wang, and Darrell]{xu2020hierarchical}
Jingwei Xu, Huazhe Xu, Bingbing Ni, Xiaokang Yang, Xiaolong Wang, and Trevor Darrell.
\newblock Hierarchical style-based networks for motion synthesis.
\newblock In \emph{ECCV}, 2020.

\bibitem[Xu et~al.(2023)Xu, Li, Wang, and Gui]{xu2023interdiff}
Sirui Xu, Zhengyuan Li, Yu-Xiong Wang, and Liang-Yan Gui.
\newblock Interdiff: Generating 3d human-object interactions with physics-informed diffusion.
\newblock In \emph{ICCV}, 2023.

\bibitem[Xu et~al.(2024)Xu, Wang, Wang, and Gui]{xu2024interdreamer}
Sirui Xu, Ziyin Wang, Yu-Xiong Wang, and Liang-Yan Gui.
\newblock Interdreamer: Zero-shot text to 3d dynamic human-object interaction.
\newblock \emph{arXiv preprint arXiv:2403.19652}, 2024.

\bibitem[Xue et~al.(2023)Xue, Gao, Xing, Mart{\'\i}n-Mart{\'\i}n, Wu, Xiong, Xu, Niebles, and Savarese]{xue2023ulip}
Le Xue, Mingfei Gao, Chen Xing, Roberto Mart{\'\i}n-Mart{\'\i}n, Jiajun Wu, Caiming Xiong, Ran Xu, Juan~Carlos Niebles, and Silvio Savarese.
\newblock Ulip: Learning a unified representation of language, images, and point clouds for 3d understanding.
\newblock In \emph{Proceedings of the IEEE/CVF conference on computer vision and pattern recognition}, pages 1179--1189, 2023.

\bibitem[Yang et~al.(2023)Yang, Liu, Zhang, Pan, Guo, Li, Chen, Gao, Guo, and Zhang]{yang2023lidar}
Senqiao Yang, Jiaming Liu, Ray Zhang, Mingjie Pan, Ziyu Guo, Xiaoqi Li, Zehui Chen, Peng Gao, Yandong Guo, and Shanghang Zhang.
\newblock Lidar-llm: Exploring the potential of large language models for 3d lidar understanding.
\newblock \emph{AAAI 2025}, 2023.

\bibitem[Yuan et~al.(2023)Yuan, Song, Iqbal, Vahdat, and Kautz]{yuan2023physdiff}
Ye Yuan, Jiaming Song, Umar Iqbal, Arash Vahdat, and Jan Kautz.
\newblock Physdiff: Physics-guided human motion diffusion model.
\newblock In \emph{ICCV}, 2023.

\bibitem[Zhang et~al.(2023)Zhang, Rao, and Agrawala]{zhang2023adding}
Lvmin Zhang, Anyi Rao, and Maneesh Agrawala.
\newblock Adding conditional control to text-to-image diffusion models.
\newblock In \emph{Proceedings of the IEEE/CVF international conference on computer vision}, pages 3836--3847, 2023.

\bibitem[Zhang et~al.(2024{\natexlab{a}})Zhang, Cai, Pan, Hong, Guo, Yang, and Liu]{zhang2024motiondiffuse}
Mingyuan Zhang, Zhongang Cai, Liang Pan, Fangzhou Hong, Xinying Guo, Lei Yang, and Ziwei Liu.
\newblock Motiondiffuse: Text-driven human motion generation with diffusion model.
\newblock \emph{PAMI}, 2024{\natexlab{a}}.

\bibitem[Zhang et~al.(2022{\natexlab{a}})Zhang, Guo, Zhang, Li, Miao, Cui, Qiao, Gao, and Li]{zhang2022pointclip}
Renrui Zhang, Ziyu Guo, Wei Zhang, Kunchang Li, Xupeng Miao, Bin Cui, Yu Qiao, Peng Gao, and Hongsheng Li.
\newblock Pointclip: Point cloud understanding by clip.
\newblock In \emph{CVPR 2022}, 2022{\natexlab{a}}.

\bibitem[Zhang et~al.(2022{\natexlab{b}})Zhang, Zhang, Fang, Gao, Li, Dai, Qiao, and Li]{zhang2022tip}
Renrui Zhang, Wei Zhang, Rongyao Fang, Peng Gao, Kunchang Li, Jifeng Dai, Yu Qiao, and Hongsheng Li.
\newblock Tip-adapter: Training-free adaption of clip for few-shot classification.
\newblock In \emph{ECCV 2022}. Springer Nature Switzerland, 2022{\natexlab{b}}.

\bibitem[Zhang et~al.(2024{\natexlab{b}})Zhang, Han, Liu, Zhou, Lu, Qiao, Li, and Gao]{zhang2024llama}
Renrui Zhang, Jiaming Han, Chris Liu, Aojun Zhou, Pan Lu, Yu Qiao, Hongsheng Li, and Peng Gao.
\newblock Llama-adapter: Efficient fine-tuning of large language models with zero-initialized attention.
\newblock In \emph{ICLR 2024}, 2024{\natexlab{b}}.

\bibitem[Zhang et~al.(2024{\natexlab{c}})Zhang, Jiang, Zhang, Lin, Guo, Qiu, Zhou, Lu, Chang, Gao, et~al.]{zhang2024mathverse}
Renrui Zhang, Dongzhi Jiang, Yichi Zhang, Haokun Lin, Ziyu Guo, Pengshuo Qiu, Aojun Zhou, Pan Lu, Kai-Wei Chang, Peng Gao, et~al.
\newblock Mathverse: Does your multi-modal llm truly see the diagrams in visual math problems?
\newblock \emph{ECCV 2024}, 2024{\natexlab{c}}.

\bibitem[Zhang et~al.(2024{\natexlab{d}})Zhang, Wei, Jiang, Guo, Li, Zhang, Tong, Liu, Zhou, Wei, et~al.]{zhang2024mavis}
Renrui Zhang, Xinyu Wei, Dongzhi Jiang, Ziyu Guo, Shicheng Li, Yichi Zhang, Chengzhuo Tong, Jiaming Liu, Aojun Zhou, Bin Wei, et~al.
\newblock Mavis: Mathematical visual instruction tuning with an automatic data engine.
\newblock \emph{arXiv preprint arXiv:2407.08739}, 2024{\natexlab{d}}.

\bibitem[Zhong et~al.(2024)Zhong, Xie, Jampani, Sun, and Jiang]{zhong2025smoodi}
Lei Zhong, Yiming Xie, Varun Jampani, Deqing Sun, and Huaizu Jiang.
\newblock Smoodi: Stylized motion diffusion model.
\newblock In \emph{ECCV}, 2024.

\bibitem[Zhu et~al.(2023)Zhu, Lin, Ning, Yan, Cui, Wang, Pang, Jiang, Zhang, Li, et~al.]{zhu2023languagebind}
Bin Zhu, Bin Lin, Munan Ning, Yang Yan, Jiaxi Cui, HongFa Wang, Yatian Pang, Wenhao Jiang, Junwu Zhang, Zongwei Li, et~al.
\newblock Languagebind: Extending video-language pretraining to n-modality by language-based semantic alignment.
\newblock \emph{arXiv preprint arXiv:2310.01852}, 2023.

\bibitem[Zhu* et~al.(2023)Zhu*, Zhang*\#, He, Guo, Zeng, Qin, Zhang, and Gao]{zhu2023pointclip}
Xiangyang Zhu*, Renrui Zhang*\#, Bowei He, Ziyu Guo, Ziyao Zeng, Zipeng Qin, Shanghang Zhang, and Peng Gao.
\newblock Pointclip v2: Prompting clip and gpt for powerful 3d open-world learning.
\newblock \emph{ICCV 2023}, 2023.

\end{thebibliography}
}


\end{document}